%% file: main.tex
\documentclass[10pt,twocolumn,letterpaper]{article}

\usepackage[pagenumbers]{cvpr} 

\input{preamble}

\definecolor{cvprblue}{rgb}{0.21,0.49,0.74}
\usepackage[pagebackref,breaklinks,colorlinks,allcolors=cvprblue]{hyperref}

\title{XTrack: Multimodal Training Boosts RGB-X Video Object Trackers}
\usepackage{multirow}
\usepackage{makecell}
\usepackage{supertabular}
\usepackage{colortbl}
\usepackage{svg}

\usepackage[dvipsnames]{xcolor}

\author{Yuedong Tan$^{1}$\quad Zongwei Wu$^{1}$\quad Yuqian Fu$^{2}$ \quad  Zhuyun Zhou$^{1}$ \quad Guolei Sun$^{4}$
    \quad Eduard Zamfir $^{1}$   \\  Chao Ma$^{3}$  \quad Danda Pani Paudel$^{2}$ \quad Luc Van Gool$^{2,4}$ \quad Radu Timofte$^{1}$
    \\ \small
     $^{1}$ Computer Vision Lab, CAIDAS \& IFI, University of Wurzburg  \quad 
        $^2$ INSAIT, Sofia University \\ 
         \small $^3$ AI Institute, Shanghai Jiao Tong University    \quad     $^4$  CVL, ETH Zurich 
}

\begin{document}
\maketitle

\input{sec/0_abstract}    
\input{sec/1_newIntro}

\input{sec/2_related_work}

\input{sec/3_method}

\input{sec/4_experiments}

\input{sec/5_discussion}

{
    \small
    \bibliographystyle{ieeenat_fullname}
    \bibliography{main}
}


\end{document}

%% file: preamble.tex
\usepackage{amsmath} 
\usepackage{multirow} 
\usepackage{booktabs}
\usepackage{colortbl}
\usepackage{makecell}
\usepackage{pifont}
\usepackage{dblfloatfix}
\usepackage{tabularx}
\usepackage{placeins}
\usepackage{caption}
\colorlet{lightorange}{Dandelion!50}
\colorlet{lightyellow}{Yellow!30}
\colorlet{rowhighlight}{Blue!13}

%% file: sec/0_abstract.tex
\begin{abstract}

Multimodal sensing has proven valuable for visual tracking, as different sensor types offer unique strengths in handling one specific challenging scene where object appearance varies.  While a generalist model capable of leveraging all modalities would be ideal, development is hindered by data sparsity, typically in practice, only one modality is available at a time. Therefore, it is crucial to ensure and achieve that knowledge gained from multimodal sensing -- such as identifying relevant features and regions -- is effectively shared, even when certain modalities are unavailable at inference.  We venture with a simple assumption: similar samples across different modalities have more knowledge to share than otherwise. To implement this, we employ a ``weak" classifier tasked with distinguishing between modalities. More specifically, if the classifier ``fails" to accurately identify the modality of the given sample, this signals an opportunity for cross-modal knowledge sharing. Intuitively, knowledge transfer is facilitated whenever a sample from one modality is sufficiently close and aligned with another. Technically, we achieve this by routing samples from one modality to the expert of the others, within a mixture-of-experts framework designed for multimodal video object tracking. During the inference,  the expert of the respective modality is chosen, which we show to benefit from the multimodal knowledge available during training, thanks to the proposed method. Through the exhaustive experiments that use only paired RGB-E, RGB-D, and RGB-T during training, we showcase the benefit of the proposed method for RGB-X tracker during inference, with an average +3\% precision improvement over the current SOTA. Our source code is publicly available \href{https://github.com/supertyd/XTrack/tree/main}{here}.  

\end{abstract} 

%% file: sec/1_newIntro.tex
\section{Introduction}
\label{sec:intro}

\begin{figure}[t]
\centering
\includegraphics[width=\linewidth,keepaspectratio]{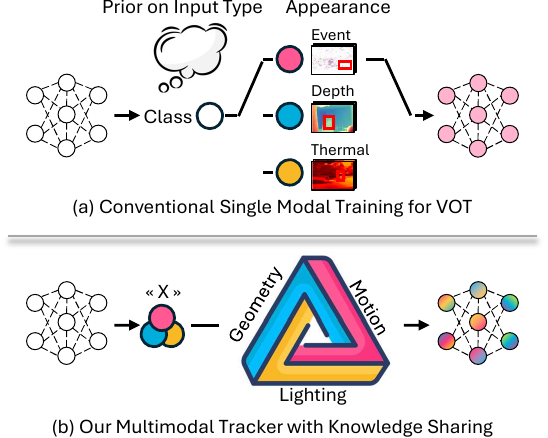}
\vspace{-5mm}
\caption{\textbf{Motivation:} (a) Existing tracking methods typically address each modality in isolation, tackling one appearance-related challenge at a time. This is mainly due to cross-modal domain gaps and the lack of a comprehensive multimodal dataset.
Consequently, only modality-specific branches are activated during inference based on a priori knowledge of the input type, limiting the potential for cross-modal integration. (b) We make this cross-modal ``impossible triangle" possible by decomposing the knowledge from each modality into transferable attributes, each capturing distinct environmental aspects, which is achieved for the first time for an RGB-X tracker.}
\vspace{-4mm}
\label{fig:teaser}
\end{figure}

Visual object tracking has made significant progress, driven by recent advancements in deep learning \cite{siamban,transt,seqtrack}. While RGB-based trackers have demonstrated strong, generalizable performance \cite{seqtrack,artrack,swintrack,sparsett}, they still struggle with large appearance changes under challenging scenes \cite{lasot,got10k,trackingnet}. Additional sensory input can enhance robustness in these scenarios, with each sensor addressing a specific challenge — such as Event cameras\cite{fe108,wang2024event} for motion awareness, depth sensors\cite{rgbd1k} for occlusion handling, and thermal imaging\cite{gtot} for illumination variation — leading to increasing interest in multimodal object tracking \cite{depthtrack,li2021lasher,vipt}.

However, current advances in multimodal tracking are limited by the lack of a comprehensive dataset that includes all modalities simultaneously. Most existing datasets treat each modality separately \cite{rgbd1k,rgbt234,vot2022}. Consequently, there is significant research interest in developing multimodal tracking systems that can benefit from multimodal training using only paired data, while generalizing to any modality available during inference, achieving the same level of generalization as conventional RGB trackers \cite{vipt,hou2024sdstrack,hong2024onetracker}.

Unfortunately, these practical concerns have been largely overlooked in previous works. Early multimodal tracking approaches \cite{depthtrack,rgbd1k,li2021lasher,rgbt234} were designed to handle only one additional modality, such as Event \cite{fe108,wang2023visevent}, Depth \cite{depthtrack}, or Thermal data \cite{gtot}, making it inherently impossible to benefit from broader multimodal training. While recent works\cite{hou2024sdstrack,hong2024onetracker} have attempted to transcend modality-specific architectures by creating general processing blocks, they typically rely on predetermined branches activated based on the input modality type, assuming prior knowledge of the modality during inference \cite{untrack,SIPHOT}. This rigid separation of modality-specific parameters, while achieving unification, prevents any meaningful interaction between modalities during training. Such practice of strict modality isolation fails to capitalize on a crucial opportunity: the potential for cross-modal knowledge transfer \cite{limoe}. Consider a complex RGB-Depth sequence involving fast-moving objects under low-lighting conditions – a model should naturally acquire knowledge that transcends modality-specific boundaries, developing insights transferable across different sensing domains.

In this work, we present the first systematic approach to enable cross-modal knowledge sharing in RGB-X video object tracking. Our key insight is that similar samples across different modalities naturally share more transferable knowledge. More importantly, when feature similarity creates confusion between modalities, we posit that this ``confusion" signals an optimal opportunity for knowledge sharing, as it indicates minimal domain gap and maximal communication between modality-specific attributes.

To implement this insight, we introduce a novel approach using a ``weak" classifier to identify such ``confusion", or more properly speaking, knowledge-sharing opportunities. We realize this through a carefully designed Mixture of Modal Experts (MeME) framework. Different from previous mixture-of-experts (MoE) work aiming for inference speedup \cite{dai2024deepseekmoe,llama-moe}, we combine the MoE sparse architecture with modality-specific processing. More importantly, we design a \textit{soft} router where experts originally trained for one modality can be leveraged by another when their features align sufficiently. Specifically, we first learn a dynamic routing function that acts as a classifier with explicit modal awareness. Based on the router's soft predictions, the input modality is directed to relevant experts for specialized analysis, creating a blended representation that incorporates both shared and modality-specific knowledge.

At the expert level, our MeME framework employs a shared unit for general knowledge alongside several modality-specific experts focusing on distinct attributes (e.g., geometry, motion, low-light adaptation). This attribute-based, flexible understanding enables comprehensive cross-modal intelligence, advancing our tracker's capability to handle complex scenarios without requiring prior modality information, which is the first of its kind for VOT.

Through extensive experiments with paired RGB-E \cite{wang2023visevent}, RGB-D \cite{depthtrack}, and RGB-T \cite{li2021lasher} data during training, we show that our approach successfully preserves and transfers multimodal knowledge to enhance single-modality tracking during inference. Our method achieves an average precision improvement of 3 points over the current state-of-the-art across various RGB-X tracking scenarios, establishing a new paradigm for robust multimodal tracking in real-world conditions where sensor availability cannot be guaranteed.

\begin{figure}[t]
\centering
\includegraphics[width=\linewidth,keepaspectratio]{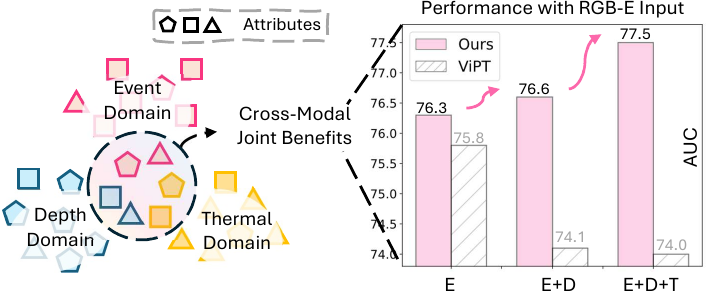}
\vspace{-7mm}
\caption{\textbf{Joint Benefits:} Despite domain gaps, some samples across modalities share similar attributes, creating overlap in representation, as shown in the left. This overlap complicates strict modality classification, introducing ambiguity. For example, on the Event benchmark, SOTA methods like ViPT \cite{vipt} perform worse when trained on multiple modalities than on Events alone. In contrast, we view this ambiguity as a chance for cross-modal knowledge sharing. Our approach leverages this potential, enabling effective multimodal training and consistent improvement.
}
\vspace{-3mm}
\label{fig:teaser2}
\end{figure}

%% file: sec/2_related_work.tex
\section{Related Work}
\label{sec:related_work}

\textbf{Multi-Modal Tracking:} 
In the evolution of object tracking \cite{zhao2023representation,atom,siamban,bacf}, the model architecture has undergone a continuous transformation, transitioning from correlation filters \cite{bacf,kcf} to Siamese networks \cite{siambag,siamcar,siamfc}, and then to the current transformer architecture \cite{transt,swintrans} that integrates unified feature extraction and interaction. However, the performance and stability of object tracking are still limited when dealing with complex scenes. 

Multi-modal information, including depth \cite{he2021fast,yang2022rgbd}, event \cite{zhang2022spiking,zhu2023cross}, and thermal data \cite{zhang2022visible,zhao2021unified}, can compensate for this deficiency and enhance the robustness of object tracking networks when dealing with objects with large appearance variation. For example, BAT \cite{bat} and TBSI \cite{tbsi} have designed a dual-transformer architecture specifically for RGBT object tracking, while DepthTrack \cite{depthtrack} is tailored for RGBD object tracking. As can be seen, most of these works are modality-tailored, without being able to transfer the knowledge from one modal domain to another. Recently, there has been a growing interest in developing multi-modal object tracking, especially following the design paradigm of prompt learning, aiming to establish a unified yet efficient object tracking paradigm \cite{vipt,hou2024sdstrack,hong2024onetracker,untrack,protrack,untrack}. However, these works adopt a rigid design, only activating the modality-specific parameter during inference, making it impossible to leverage joint benefits across all the modalities.

\noindent\textbf{Emergent Intelligence:} Previous works have demonstrated the benefits of training on diverse modalities, including images, text, and 3D data, while others have highlighted the advantages of learning across different tasks. The ultimate goal is to create a single network capable of handling any type of input, an ability inherent to humans. Some researchers refer to this as a generalist model. For example, the Meta-Transformer \cite{zhang2023meta} aligns various modalities into a unified feature space, while ImageBind \cite{girdhar2023imagebind} uses images to align data from different modalities. PolyViT \cite{likhosherstov2022polyvit} has pioneered unified training of a Vision Transformer (ViT) backbone with multiple modalities.

These approaches highlight the importance of a joint learning space, where knowledge can be transferred between domains. In this paper, we aim to achieve this goal within the context of multimodal object tracking, addressing this challenge for the first time.

\noindent\textbf{Mixture-of-Experts:}
Mixture-of-Experts (MoE) has emerged as a powerful technique that enhances model capability and efficiency by delegating tasks to specialized expert networks. This architecture is particularly effective for handling complex and diverse data, as it dynamically selects the most appropriate expert based on the input's characteristics. Models such as VMoE \cite{vmoe} and DAMeX \cite{jain2024damex} have significantly improved image classification and object detection tasks by extending feed-forward network (FFN) layers within the vision transformer architecture. These models adjust the combination of their internal expert networks based on input features, thereby capturing complex patterns more effectively. DeepSeekMoE \cite{dai2024deepseekmoe} exemplifies the use of MoE in large language models, improving language understanding and generation through the expert mixture mechanism. Other models like IMP \cite{imp} and LiMoE \cite{limoe} apply MoE to unify tasks across different modalities, such as visual, auditory, and textual data processing.

However, existing works primarily associate MoE’s sparsity with task-specific needs, overlooking the joint benefits of cross-modal integration due to rigid task or modality-specific isolation. In this paper, we aim to introduce a more flexible and intelligent separation strategy that facilitates cross-modal knowledge transfer for VOT.

%% file: sec/3_method.tex
\section{Method}
\label{headings}

\subsection{Overall Pipeline} 
\begin{figure*}[t]
\centering
\includegraphics[width=.99\linewidth,keepaspectratio]{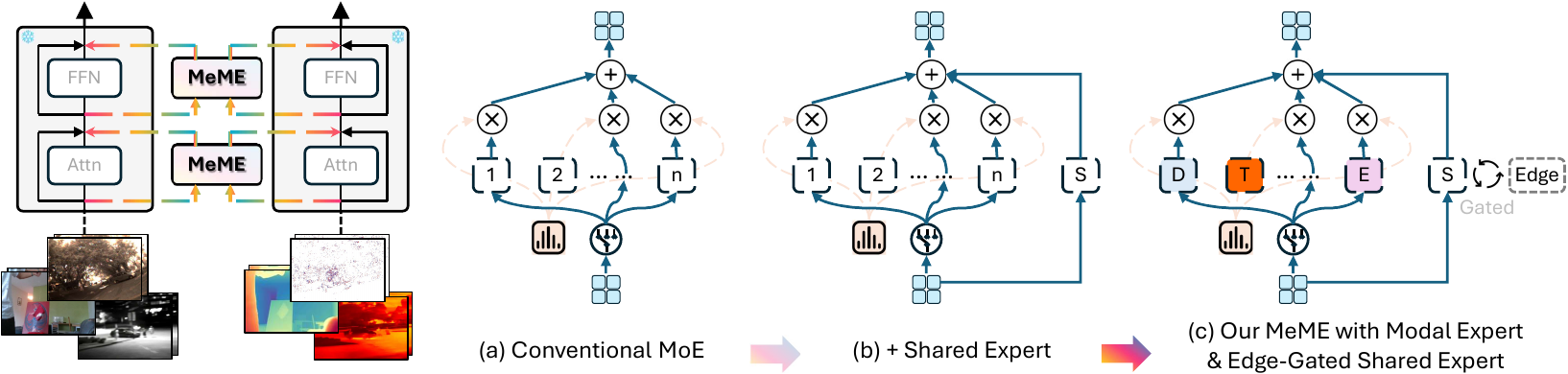}
\vspace{-2mm}
\caption{\textbf{Mixture of Modal Experts (MeME):} MoE has demonstrated significant advancements recently. Recent work \cite{dai2024deepseekmoe} has further leveraged shared experts to reduce redundancy. However, it remains unclear which expert learns what, due to the implicit learning setting. In contrast, we make the learning protocol explicit by assigning specific experts to particular modal inputs. Additionally, we enhance the shared expert model with inductive edge bias, increasing efficiency when dealing with relatively limited downstream data.
}
\vspace{-3mm}
\label{fig:meme}
\end{figure*}

In line with the current tracking paradigm \cite{vipt,hong2024onetracker}, we input both the template and the search region, along with their respective multi-modal patches, into the same patch embedding to obtain the RGB tokens $T_{rgb}$ and the auxiliary modal tokens $T_{x}$. To ensure effective interaction between the two modalities, as illustrated in Fig. \ref{fig:meme}(left), we incorporate the \textbf{Mixture of Modal Experts (MeME)}, which enables intertwining between RGB and X input. Our MeME takes place after each attention block (Attn) and the Feed-Forward Network (FFN) of our baseline foundation RGB trackers, built using transformers with frozen weights \cite{coco,lasot,trackingnet,got10k}.

Our goal is to leverage the joint benefits of the RGB and X modalities to enhance feature modeling and better handle large appearance variations across video frames. To achieve this, MeME is designed in a bidirectional manner, allowing the X modality to improve the RGB features, and vice versa. For simplicity and clarity, we show in the following section how the modal token $T_{x}$ improves the RGB token $T_{rgb}$.

As shown in Fig. \ref{fig:meme}(left), at the $l^{th}$ transformer block, our MeME takes the RGB token $T_{rgb}^{l}$ and the X token $T_{x}^{l}$ from the previous transformer layer outputs. We first merge these two feature sets and enhance the conventional feature modeling with transformer attention:
\begin{equation}
\label{eq:attn}
T_{rgb}^{attn} = T_{rgb}^{l} + Attn(T_{rgb}^{l}) + MeME(T_{rgb}^{l}, T_{x}^{l}).
\end{equation}

Similarly, we enable the cross-modal interaction again to complement the FFN processing. We have:
\begin{equation}
T_{rgb}^{l+1} = T_{rgb}^{attn} + FFN(T_{rgb}^{attn}) + MeME(T_{rgb}^{attn}, T_{x}^{attn}),
\end{equation}
where $T_{x}^{attn}$ is computed in the similar way as in Eq. \ref{eq:attn}.

\subsection{Mixture of Modal Experts (MeME)}
\label{sec:meme}

 MeME operates on two key principles: (a) assigning the most appropriate and specialized experts based on the input modality, and (b) isolating the shared experts that are common across all modal inputs to facilitate cross-modal alignment and reasoning. By jointly incorporating both specialized and shared experts, MeME allows for the exploration of modal-specific clues with finer granularity thanks to the reduced redundancy within each specialized expert, and ensures emergent alignment across modalities, even when trained solely on paired data.

\label{sec:loss}

\noindent \textbf{Router Supervision:} One key component of an MoE architecture is the routing function:
\begin{equation}
    y=\sum_{i \in   top\_k} p_i(T_{x}) \epsilon_i(T_{x}),
\end{equation}
where we adopt $top\_k$ experts for routing the input, $p_i$ is the probability, and $\epsilon_i$ is the expert. We supervise such a routing function by two losses: the widely adopted expert balance loss and our newly introduced classification loss:

\noindent \underline{\textit{Expert Balance Loss:}}  Balance loss is composed of two parts, the importance loss $\mathcal{L}_{Imp}$ and the loading loss $\mathcal{L}_{Load}$. Let $I$ denote the vector of importance scores for all experts, where each element $I_i$ represents the importance score of expert $\epsilon_i$. The importance score $I_i$ is calculated as the sum of the probabilities $p_i$ over all samples. 
 We consider a norm distribution denoted as $\mathcal{N}(0,\sigma^2\textit{I})$, where the standard deviation $\sigma$ is defined as the ratio of gate noise to the number of experts $|\epsilon|$. $\Phi(\cdot)$ is the cumulative distribution function (CDF) of this normal distribution.

For the $Load_i$ of expert $e_i$, we evaluate the CDF at the probabilities $p_i(T_{x})$, represented as $\Phi(p_i(T_{x}))$, and sum these values over all inputs $T_{x}$ in a single mini-batch. The load loss $\mathcal{L}_{Load}$ is then defined as:

\begin{equation}
    \mathcal{L}_{Load} = \frac{\operatorname{Var}(Load_i)}{\operatorname{Mean}(Load_i)^2}; \; Load_i = \sum_{T_{x} \in B}\Phi(p_i(T_{x})).
\end{equation}
Here, $Load_i$ represents the number of assignments allocated to expert $e_i$, and $B$ stands for the batch. Similarly, we obtain the importance loss $\mathcal{L}_{Imp}$, but without considering the CDF, by:
\begin{equation}
    \mathcal{L}_{Imp} = \frac{\operatorname{Var}(Imp_i)}{\operatorname{Mean}(Imp_i)^2}; \; Imp_i = \sum_{T_{x} \in B} p_i(T_{x}).
\end{equation}

Hence, our balance loss $\mathcal{L}_{balance}$ is formulated by:
\begin{equation}
    \mathcal{L}_{balance} = \mathcal{L}_{Imp}+\mathcal{L}_{Load}.
\end{equation}

\noindent \underline{\textit{Classification Loss:}}   To enable each expert in our MeME to benefit from multimodal information while maintaining a certain degree of modality specialization, we introduced modality classification loss on the basis of ensuring a balanced expert load. This allows the model to learn extensive expert knowledge from various modalities and absorb some knowledge from other modalities as well. The router, while maintaining the passage through its modality-specific experts, also leverages experts from other modalities.

To achieve this soft routing, we additionally employ a multi-class loss to constrain the data that each expert can process. Given a collection $M$ of modalities, indexed by $n$, denoted as ${m_n}_{n=1}^{|M|}$. We introduce a mapping function $h: M \rightarrow \epsilon$ such that each modality $m_n$ is associated with several particular experts $e_i \in \epsilon$, which ideally are not shared from one modality to another. The classification loss $\mathcal{L}_{cls}$ is computed using a binary cross-entropy loss, comparing the logits $p_i$ (representing the probability of selecting experts) with the labels $h(m_n)$ (indicating the target expert for tokens from modality $m_n$). The expression for $\mathcal{L}_{cls}$ is formulated as follows:
\begin{equation}
\mathcal{L}_{cls}=-\sum_{n=1}^{|M|} \sum_{i=1}^{|\epsilon|} 1_{\left\{h\left(m_n\right)=i\right\}} \log p_i(T_{x}),
\end{equation}
where $p_i(T_{x})$ is the probability,  $1_{\{h(m_n) = i\}}$ is the indicator function that equals to 1 if $h(m_n) = i$ and 0 otherwise, and $|\epsilon|$ is the cardinality of the set of experts $\epsilon$.

Finally, during training and on top of conventional tracking, our model additionally leverages MoE loss $\mathcal{L}_{moe}$ to achieve our targeted soft routing setting:

\begin{equation}
    \mathcal{L}_{moe} = \mathcal{L}_{cls} + \lambda \cdot \mathcal{L}_{balance},
\end{equation}
where $\lambda$ is the proportion hyperparameter.

\subsection{Efficient Expertized Reasoning and Gathering}
\textbf{Experts with Low-Dimensional Reasoning:} 
The goal of each specialized expert is to project the input feature into a dedicated space for further processing. Inspired by recent successes in tuning large language models \cite{llama-moe,wang2024visionllm}, we aim to enhance efficiency by mitigating low-dimensional feature decomposition and reconstruction.  Specifically, we project each input token with channel $c$ to the significantly lower dimensional latent space $k$ ($k << c$). This approach preserves the principal features while reducing computational cost, making the learning feasible on a single 24 GB GPU.

Assume that our router decomposes the mixed modal input $M$ into subset features $m_n$. Their respective $k_{th}$ low-dimensional matrices ${m_n}_k$, are approximated by:
\begin{equation}
    {m_n}_k = \epsilon_n(m_n),
\end{equation}
where $\epsilon_n$ mimics the lower dimension projection for key feature transformation. Note that all the $\epsilon_n$ structures are identical for each modality-specific expert but without weight sharing, allowing for distinct reasoning.

\vspace{1mm}

\noindent \textbf{Edge-Gated Shared Expert:} Simultaneously, we compute the shared low-dimensional matrix $m_s$ from the shared expert $\epsilon_s$. However, we believe that for the shared expert, a single projection through implicit learning might not be sufficient, especially when the training data size is relatively small compared to that used for training large foundation models. Therefore, we think it is necessary to incorporate additional human prior to guide the shared learning process. Since the objective is to find commonalities, we believe that high frequency is a naturally shared feature among these modalities, typically manifesting in the form of edges.

\begin{figure}[!t]
\centering
\includegraphics[scale=0.65]{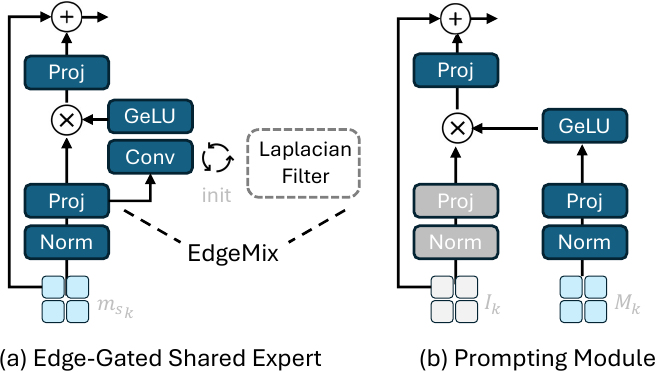}
\vspace{-2mm}
\caption{Details on experts and prompts.
}
\vspace{-3mm}
\label{fig:blocks}
\end{figure}

Technically, we introduce a gating module at low latent space.  A graphical illustration of this process can be found in Fig. \ref{fig:blocks}(a). Let ${m_s}_k$ be the input shared low-dimension feature, we obtain the output $Out$ by:
\begin{equation}
\begin{split}
    & X = Norm({m_s}_k), \\
    & Out = (\sigma(EdgeMix(X W_1)) \cdot X W_2) W_3 + {m_s}_k,
\end{split}
\label{eq:edgemix}
\end{equation}
where $Norm()$ denotes the conventional normalization. $EdgeMix()$ refers to the module, with Laplacian initialization, responsible for conducting token mixing, as shown in Fig. \ref{fig:blocks}(a). Technically, we begin by transferring the token to a 2D feature form and then apply convolution to enable interaction between neighboring tokens. Unlike previous approaches, in our case, we additionally initialize the convolution with a Laplacian Filter, incorporating edge priors while remaining open to discovering more meaningful and learnable shared features. $W_1, W_2, W_3$ represent learnable parameters within the low-dimensional latent space.

\vspace{1mm}

\noindent \textbf{Fusion of Expertized Tokens:} The next step aims to fuse previously separately reasoned tokens together into a global low-dimensional matrix $M_k$. It should contain both modality-specific and modality-shared clues. Technically, we develop a shrinkage fusion design, where we first concatenate ${m_n}_k$ together batch-wise. Then, we learn a joint approximation from the whole batch, which is further incorporated with the shared matrix $m_s$. This pipeline can be expressed as follows: 
\begin{equation}
M_k = ([{m_n}_k] W_4 + {m_s}_k) W_5,
\label{eq:overall}
\end{equation}
where $[.]$ is the batch combination and $W_4, W_5$ are low latent space learnable parameters.

\subsection{Modal Prompting}

Finally, we aim to prompt the RGB token obtained from the frozen RGB foundation tracker and transform them to be modality-aware, making them more suitable for downstream and challenging cases. Following the same idea, we first project the RGB token into the low latent space $I_k$. Then, we use approximated modal low-dimensional matrix $M_k$ from Eq. \ref{eq:overall}  to prompt $I_k$. Our motivation is to use the modality clues as a gating module to improve the RGB feature modeling. Specifically, we have: 
\begin{equation}
\begin{split}
    & X_i = Norm({I}_k), \, \; X_m = Norm({M}_k), \\
    & Out = ((X_i W_5 \cdot \sigma(X_m W_6)) W_7 + {I}_k) W_8,
\end{split}
\label{eq:tokenmix}
\end{equation}
where $W_5, W_6, W_7$ are the learning parameters in the low latent space, and $W_8$ projects the prompted token back to the initial embedding space.

%% file: sec/4_experiments.tex
\section{Experiments}

We adopt \cite{ostrack} and \cite{wu2023dropmae} as RGB foundation trackers by freezing the parameters, forming XTrack-B and XTrack-L, respectively. To train our XTrack, only the parameters in MeME are learnable, as shown in Fig. \ref{fig:meme}. We set the batch size to 32, with a learning rate of 4e-4, training for 90 epochs. The learning rate is decreased by a factor of 10 after 78 epochs. In addition to the object tracking loss shared with the RGB Tracker, we incorporate a loss that guides MoE expert balance and specialization of modality-specific experts, as detailed in Sec. \ref{sec:loss}.

Regarding dataset selection, we utilize DepthTrack \cite{depthtrack} for RGB-Depth sequences, LasHeR \cite{li2021lasher} for RGB-Thermal sequences, and VisEvent \cite{wang2023visevent} for RGB-Event sequences. Previous well-established modality-tailored works \cite{vipt,protrack,hong2024onetracker,hou2024sdstrack}, we train our model from all available modalities. We want to highlight that only one RGB-X pair (i.e. RGB-Depth, \textbf{\textit{or}} RGB-Thermal, \textbf{\textit{or}} RGB-Event) is available at a time, which makes our problem setting of knowledge transfer challenging.

\subsection{Depth Benchmark}

We compare our XTrack with two groups of competitors: 1) specific models only for tackling depth dataset, e.g., SPT~\cite{spt}, DeT~\cite{depthtrack}, DDiMP~\cite{vot2020}, ATCAIS~\cite{vot2020}, MaCNet~\cite{macnet}, DAFNet~\cite{dafnet}, and so on; 2) SOTA generalizable models but with tailored depth parameters, e.g., ProTrack~\cite{prompt}, ViPT~\cite{vipt}, UnTrack~\cite{untrack}, OneTracker \cite{hong2024onetracker}, and SDSTrack \cite{hou2024sdstrack}.

\noindent \textbf{DepthTrack:} DepthTrack~\cite{depthtrack} is a comprehensive, long-distance dataset consisting of 50 video sequences for test. We evaluate the accuracy of various trackers on this dataset using F-score, Recall (Re), and Precision (Pr) metrics. Our results show that: 1) XTrack outperforms all models specifically designed for the DepthTrack dataset, as well as unified models using depth-specialized parameters. 2) With an increase in model scale from XTrack-B to XTrack-L, we achieve notable improvements, setting new SOTA records.

\noindent \textbf{VOT-RGBD2022:} The VOT-RGBD2022 dataset~\cite{vot2022}, which includes 127 video sequences, is the latest benchmark for RGB-D object tracking. We evaluate our model's performance using the metrics of expected average overlap (EAO), accuracy, and robustness.

Note that VOT-RGBD2022 contains out-of-distribution samples, creating a domain gap between the DepthTrack training set and the VOT-RGBD2022 test set. However, as shown in Tab.~\ref{tab:rgbd}, our method surpasses current state-of-the-art performance on VOT-RGBD2022 by an even larger margin than on DepthTrack. This result highlights our model's enhanced capability to manage domain gaps, a strength we attribute to our multimodal training.

\begin{table}[t]
    \caption{Performance on RGB-Depth datasets.
    }
    \vspace{-5mm}
\label{tab-sota-rgbd}
\vspace{2mm}
  \centering
\resizebox{1\linewidth}{!}{
  \setlength{\tabcolsep}{1.5mm}{
    \small
    \begin{tabular}{l|ccc c ccc}
    \toprule
    \multirow{2}*{Method} & \multicolumn{3}{c}{DepthTrack~\cite{depthtrack}} & & \multicolumn{3}{c}{VOT-RGBD22~\cite{vot2022}} \\
        \cline{2-4} \cline{6-8}
 & F-score & Re & Pr& & EAO & Acc.& Rob. \\
    \midrule[0.5pt]

ATOM~\cite{atom} &-&-&- & &50.5&59.8&68.8\\
DiMP~\cite{dimp} &-&-&- & &54.3&70.3&73.1\\
CA3DMS~\cite{ca3dms} &22.3&22.8&21.8 & &-&-&-\\
SiamM-Ds~\cite{VOT2019} &33.6&26.4&46.3 & &-&-&-\\
Siam-LTD~\cite{vot2020} &37.6&34.2&41.8 & &-&-&-\\
LTDSEd~\cite{VOT2019} &40.5&38.2&43.0 & &-&-&-\\
DAL~\cite{dal} &42.9&36.9&51.2 & &-&-&-\\
GLGS-D~\cite{vot2020} &45.3&36.9&58.4 & &-&-&-\\
LTMU-B~\cite{LTMU} &46.0&41.7&51.2 & &-&-&-\\
ATCAIS~\cite{vot2020} &47.6&45.5&50.0 & &55.9&76.1&73.9\\
DRefine~\cite{vot2021} &-&-&- & &59.2&77.5&76.0\\
KeepTrack~\cite{keeptrack} &-&-&- & &60.6&75.3&79.7\\
DDiMP~\cite{vot2020} &48.5&56.9&50.3 & &-&-&-\\
DMTrack~\cite{vot2022} &-&-&- & &65.8&75.8&85.1\\
DeT~\cite{depthtrack} &53.2&50.6&56.0 & &65.7&76.0&84.5\\
OSTrack~\cite{ostrack} &52.9&52.2&53.6 & &67.6&80.3&83.3\\
SBT-RGBD~\cite{sbt} &-&-&- & &70.8&80.9&86.4\\
SPT~\cite{rgbd1k} &53.8&54.9&52.7 & &65.1&79.8&85.1\\
        \midrule[0.1pt]

\rowcolor{gray!8}
ProTrack~\cite{protrack} &57.8&57.3&58.3 & &65.1&80.1&80.2\\
\rowcolor{gray!8}
ViPT~\cite{vipt} &59.4&59.6&59.2 & &72.1&81.5 &87.1\\
\rowcolor{gray!8}
UnTrack~\cite{untrack} &61.0&60.8&61.1 & &72.1&82.0 &86.9\\
\rowcolor{gray!8}
OneTracker~\cite{hong2024onetracker} &60.9 &60.4 &60.7 & &72.7 & 81.9 & 87.2\\
\rowcolor{gray!8}
SDSTrack~\cite{hou2024sdstrack} &61.4 &60.9 & \textcolor{blue}{61.9} & &72.8 & 81.2 & 88.3\\

        \midrule[0.1pt]
        \rowcolor{gray!20}
            XTrack-B &\textcolor{blue}{61.5}&\textcolor{blue}{62.0}& 61.8 & &\textcolor{red}{\textbf{74.0}}&\textcolor{blue}{82.1} &\textcolor{blue}{88.8}\\
            \rowcolor{gray!20}
        XTrack-L &\textcolor{red}{\textbf{64.8}}&\textcolor{red}{\textbf{64.3}}&\textcolor{red}{\textbf{65.4}} & &\textcolor{red}{\textbf{74.0}}&\textbf{\textcolor{red}{82.8}} &\textcolor{red}{\textbf{88.9}}\\

    \bottomrule
    \end{tabular}
    }
  }
\label{tab:rgbd}
  \vspace{-2mm}
\end{table}

\subsection{Thermal Benchmark}

We compare our XTrack with two groups of competitors: 1) task specific model SGT~\cite{sgt}, CAT~\cite{cat}, DAFNet~\cite{dafnet}, MaCNet~\cite{macnet}, APFNet~\cite{apfnet}, and so on; 2) We compare with the same group of unified models but specialized for RGB-Thermal such as  ProTrack~\cite{prompt}, ViPT~\cite{vipt}, and so on.

\noindent \textbf{LasHeR:} The LasHeR~\cite{li2021lasher} dataset is a large-scale, high-quality dataset comprising 245 test sequences. We evaluate the performance of trackers using precision (Pr) and success plots (Sr). 
Notably, XTrack achieves SOTA on this dataset clearly outperforming all the competitors, without requiring any parameter shifting from one depth to thermal.

\noindent \textbf{RGBT234:} The RGBT234~\cite{rgbt234} dataset is a comprehensive video dataset, including 234 video sequences for testing, encompassing a variety of environmental challenges, including rainy conditions, nighttime scenes, and extreme weather scenarios. Results in Tab.~\ref{tab:rgbt} show our consistent improvement over SOTA methods across all the sequences.

\begin{table}[t]
    \caption{Performance on RGB-Thermal datasets.
    }
        \vspace{-5mm}

\label{tab-sota-rgbt}
\vspace{2mm}
  \centering
\resizebox{1\linewidth}{!}{
  \setlength{\tabcolsep}{2mm}{
    \small
    \begin{tabular}{l|ccccc}
    \toprule
    \multirow{2}*{Method} & \multicolumn{2}{c}{LasHeR~\cite{li2021lasher}} & & \multicolumn{2}{c}{RGBT234~\cite{rgbt234}} \\
        \cline{2-3} \cline{5-6}
 & Pr & Sr & &MSR &MPR \\
    \midrule[0.5pt]
SGT~\cite{sgt} &36.5&25.1 & &72.0 &47.2 \\
DAFNet~\cite{dafnet} &-&- &&79.6 &54.4 \\
FANet~\cite{fanet} &44.1&30.9 & &78.7&55.3\\
MaCNet~\cite{macnet} &-&- &&79.0 &55.4\\
HMFT~\cite{vtuav} &43.6&31.3 & &-&-\\
CAT~\cite{cat} &45.0&31.4 &&80.4 &56.1\\
DAPNet~\cite{dapnet} &43.1&31.4 & &-&-\\
mfDiMP~\cite{mfdimp} &44.7&34.3 & &64.6&42.8\\
JMMAC~\cite{jmmac} &-&- &&79.0 &57.3\\
CMPP~\cite{cmpp} &-&- &&82.3 &57.5\\
APFNet~\cite{apfnet} &50.0&36.2 &&82.7 &57.9\\
OSTrack~\cite{ostrack} &51.5&41.2 & &72.9&54.9\\

        \midrule[0.1pt]
\rowcolor{gray!8}
ProTrack~\cite{protrack} &53.8&42.0 & & 79.5& 59.9\\
\rowcolor{gray!8}
ViPT~\cite{vipt} &65.1 &52.5 & &83.5&61.7\\
\rowcolor{gray!8}
UnTrack~\cite{untrack} &64.6&51.3 & &84.2& 62.5\\
\rowcolor{gray!8}
SDSTrack~\cite{hou2024sdstrack} &66.5&53.1 & &84.8 &62.5\\
\rowcolor{gray!8}
OneTracker~\cite{hong2024onetracker} &67.2 &53.8 & & 85.7 &64.2\\

        \midrule[0.1pt]
    \rowcolor{gray!20}
        XTrack-B &\textcolor{blue}{69.1} &\textcolor{blue}{55.7} & &\textcolor{blue}{87.4}&\textcolor{blue}{64.9}\\
    \rowcolor{gray!20}
        XTrack-L &\textbf{\textcolor{red}{73.1}}&\textbf{\textcolor{red}{58.7}} & &\textcolor{red}{\textbf{87.8}} &\textcolor{red}{\textbf{65.4}}\\

    \bottomrule
    \end{tabular}
    }
  }
    \vspace{-2mm}
  \label{tab:rgbt}
\end{table}

\subsection{Event Benchmark}

\textbf{VisEvent:}
As for the RGB-Event benchmark, we also compare our model with two kinds of tracker: event-specific ones and the same group of unified models but specialized for RGB-Event tracking.

VisEvent \cite{wang2023visevent} is a large-scale dataset containing RGB images and event data for object tracking, comprising 320 test sequences. We use precision (Pr) and success plots (Sr) to evaluate the performance of different trackers. 
The quantitative results in Tab. \ref{rgbe} validate our effectiveness, achieving new SOTA records on the Event benchmark.

\begin{table}[t]\normalsize
    \caption{Performance on RGB-Event datasets.
    }
\label{tab-sota-rgbe}
\vspace{-2mm}
  \centering
\resizebox{1\linewidth}{!}{
  \setlength{\tabcolsep}{6mm}{
    \small
    \begin{tabular}{l|cc}
    \toprule
    \multirow{2}*{Method} & \multicolumn{2}{c}{VisEvent~\cite{wang2023visevent}}\\
        \cline{2-3}
 & Pr& Sr\\
    \midrule[0.5pt]

SiamMask\_E~\cite{SiamMask}&56.2 &36.9 \\
SiamBAN\_E~\cite{siamban}&59.1 &40.5 \\
ATOM\_E~\cite{atom}&60.8 &41.2 \\
VITAL\_E~\cite{VITAL} &64.9&41.5 \\
SiamCar\_E~\cite{Stark}&59.9 &42.0 \\
MDNet\_E~\cite{MDNet} &66.1&42.6 \\
STARK\_E~\cite{Stark} &61.2&44.6 \\
PrDiMP\_E~\cite{PrDiMP}&64.4 &45.3 \\
LTMU\_E~\cite{LTMU} &65.5&45.9 \\

TransT\_E~\cite{transt} &65.0 &47.4\\
SiamRCNN\_E~\cite{SiamRCNN}&65.9 &49.9 \\
OSTrack~\cite{ostrack} &69.5&53.4\\

        \midrule[0.1pt]
        \rowcolor{gray!8}
ProTrack~\cite{protrack}&63.2 &47.1\\
\rowcolor{gray!8}
ViPT~\cite{vipt}&75.8 &59.2\\
\rowcolor{gray!8}
UnTrack~\cite{untrack}&75.5 &58.9 \\
\rowcolor{gray!8}
OneTracker~\cite{hong2024onetracker}&76.7 &60.8\\
\rowcolor{gray!8}
SDSTrack~\cite{hou2024sdstrack} &76.7 &59.7\\
 \midrule[0.1pt]
\rowcolor{gray!20}
    XTrack-B &\textcolor{blue}{77.5} &\textcolor{blue}{60.9}\\
    \rowcolor{gray!20}
        XTrack-L &\textbf{\textcolor{red}{80.5}} &\textbf{\textcolor{red}{63.3}}\\

    \bottomrule
    \end{tabular}
    }
  }
  \label{rgbe}
\end{table}

%% file: sec/5_discussion.tex
\section{Ablation Studies and Discussion}

In this section, without explicit mentioning, all the experiments are reported based on our XTrack-B.

\vspace{1mm}
\noindent \textbf{Shared Experts:}
We conduct an ablation study on the shared experts. As shown in Tab.~\ref{tab:Ablation}, we found that, for DepthTrack~\cite{depthtrack}, removing the shared experts resulted in a -3.9\% decrease in F-score. Removing the shared experts might lead to redundancy in the modality-specialized experts, reducing their effectiveness and hence leading to such a performance drop. This observation joins the conclusion from the previous work \cite{dai2024deepseekmoe}.

\vspace{1mm}

\noindent \textbf{Modality-Specific Experts:}
In addition to the shared experts, our MeME also includes modality-specific experts to capture unique information from each modality. As shown in Tab.~\ref{tab:Ablation}, we remove this design and process all modalities using a single expert. In such a case, all the experts are treated equally, losing the identification such as geometry from depth, motion from event, or temperature from thermal, which is not desirable. 
This can be also proven by the declined performance across various modalities, particularly in depth. Such a performance drop indicates the effectiveness and necessity of our decomposition and the collaboration of both modality-shared and modality-specific features. This design enables the effective modeling of unique characteristics and makes the different modalities benefit from each other.

\begin{table}[t]
    \centering
    \caption{Key Component Analysis}
    \vspace{-3mm}
    \resizebox{1\columnwidth}{!}{
    \begin{tabular}{cc|ccc|cc|cc}
        \hline
         Shared & Spec. & \multicolumn{3}{c|}{DepthTrack} & \multicolumn{2}{c|}{LasHeR} & \multicolumn{2}{c}{VisEvent} \\
        \cline{3-9}

         & & F-score & Re & Pr   & Pr & Sr & Pr & Sr \\
        \hline
 
        & \checkmark  & 57.6 & 57.5 & 57.6 & 68.0 & 54.8 & 76.2 & 59.9  \\
        \checkmark & & 59.1 & 59.2 & 59.4  & 67.8 & 54.2 & 76.5 & 59.6  \\
        \checkmark & \checkmark &  \textbf{61.5} & \textbf{62.0} & \textbf{61.8}  & \textbf{69.1} & \textbf{55.7} & \textbf{77.5} & \textbf{60.9}  \\
        \hline
    \end{tabular}}
    \vspace{-1em}
    \label{tab:Ablation}
\end{table}

\begin{figure*}[t]
\centering
\includegraphics[width=.99\linewidth,keepaspectratio]{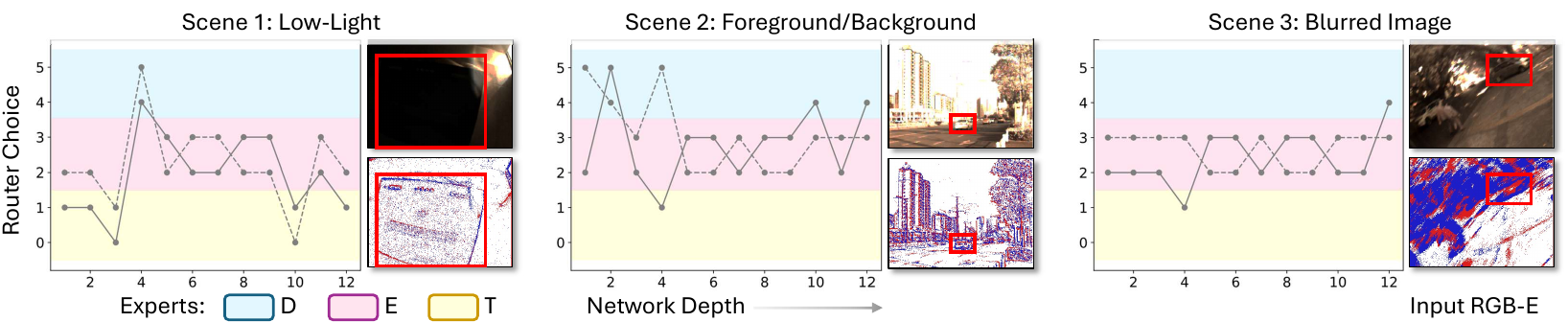}
\vspace{-2mm}
\caption{Routing Choice. We present the top-k ($k=2$) decisions for activating the \textcolor{BlueViolet}{\textbf{D}}epth, \textcolor{VioletRed}{\textbf{E}}vent, and \textcolor{Goldenrod}{\textbf{T}}hermal experts, using RGB-Event as input during inference. Our model dynamically selects the most suitable experts for different challenging scenarios, ensuring optimal object tracking performance despite appearance changes across diverse scenes.
}
\vspace{-3mm}
\label{fig:router}
\end{figure*}

\noindent \textbf{Number of Experts:} We further conduct ablations on the number of experts, by setting it to 1, 2, and 3. Experimental results in Tab. \ref{tab:Ablation1} show that using two modality-specific experts achieves the best performance. 
With only one expert per modality, feature representation is limited, while using three experts introduces internal conflicts, which also degrades model performance.

\begin{table}[t]
    \centering
        \caption{Study on the Expert Number.
    }
    \vspace{-3mm}
    \resizebox{1\columnwidth}{!}{
    \begin{tabular}{c|ccc|cc|cc}
        \hline
        \multirow{2}{*}{Number} & \multicolumn{3}{c|}{DepthTrack} & \multicolumn{2}{c|}{LasHeR} & \multicolumn{2}{c}{VisEvent} \\
        \cline{2-8}
        & F-score & Re & Pr  & Pr & Sr & Pr & Sr \\
        \hline
        1  & 60.0 & 61.2 & 60.6  & 68.8 & 54.9 & 76.0 & 59.8 \\
        2  & \textbf{61.5} & \textbf{62.0} & \textbf{61.8}  & \textbf{69.1} & \textbf{55.7} & \textbf{77.5} & \textbf{60.9} \\
        3  & 60.9 & 60.1 & 60.5  & 68.2 & 55.0 & 75.7 & 59.7 \\
        \hline
    \end{tabular}}
    \label{tab:Ablation1}
\end{table}

\vspace{1mm}

\noindent \textbf{Joint Training Benefits:} Our model is designed for cross-modal knowledge transfer, enabling it to benefit from training datasets with diverse domain samples. To verify this, we conducted experiments by progressively expanding the training set with additional modalities.

We first trained the model solely on event data and tested it on the event benchmark, mirroring previous modality-specific or unified models with event-only parameters activated. In this setting, as shown in Tab. \ref{tab:Ablation2}, our model achieved a +8\% improvement in precision over the baseline \cite{ostrack}. Interestingly, even when trained exclusively on event data, our model demonstrated strong zero-shot generalization to thermal data, achieving results comparable to ProTrack \cite{protrack}, which was trained and tested on thermal data. However, zero-shot transfer to the depth dataset proved more challenging, likely due to a greater domain gap between event and depth data than between event and thermal. This finding aligns with our model’s underlying motivation to leverage transferable knowledge across modalities: event cameras, sensitive to light changes, share similarities with thermal cameras, which also handle illumination variation effectively.

Moreover, we show that joint training with both depth and event samples boosts further our model's performance. However, we also observe that this setup does not necessarily enhance zero-shot performance on the thermal domain. This may be because the domain gap between event and depth data is too substantial. As a result, the model primarily focuses on bridging the domain gap rather than fostering cross-domain alignment. Finally, joint training with all available modalities further enhances performance across each individual modality.

\begin{table}[t]
    \centering
    \caption{Multimodal Training Benefits.}
    \vspace{-3mm}
    \resizebox{\columnwidth}{!}{
    \begin{tabular}{ccc|cc|ccc|cc}
        \hline
        \multirow{2}{*}{E} & \multirow{2}{*}{D} & \multirow{2}{*}{T} & \multicolumn{2}{c|}{VisEvent} & \multicolumn{3}{c|}{DepthTrack} & \multicolumn{2}{c}{LasHeR} \\
        \cline{4-10}
 
        & & & Pr  &  Sr & F-score & Re & Pr  & Pr & Sr \\
        \hline

            & &  &69.5  & 53.4 & 52.9 & 52.2 & 53.6  & 51.5 & 41.2\\
        \checkmark & &  &76.3  & 59.9 & 45.5 & 45.8 & 45.7  & 58.1 & 45.1\\
        \checkmark & \checkmark & &76.6  &  60.2 & 60.8 & 59.5 & 60.1 & 58.1 & 45.2 \\
        \checkmark & \checkmark & \checkmark & \textbf{77.5} &  \textbf{60.9} & \textbf{61.5} & \textbf{62.0} & \textbf{61.8} & \textbf{69.1} & \textbf{55.7} \\
        \hline
    \end{tabular}}
    \vspace{-2mm}
    \label{tab:Ablation2}
\end{table}

\vspace{1mm}

\noindent \textbf{Soft Router:}
In our router design, we use classification loss to activate the most suitable modality-specific experts. A higher classification loss pushes the classifier toward strict separation, resulting in a rigid design with parallel, non-communicating branches -- similar to current multimodal tracking models. Conversely, a lower classification loss introduces more ambiguity, approaching random expert assignment, akin to extending current unified models by treating all modalities equally without differentiation.

Tab.~\ref{tab:Ablation3} presents the performance results for the two extreme cases: rigid separation and random assignment. Our final model outperforms both, achieving the best performance with a classification probability of, amazingly, 80.00\% under the 0.001 loss proportion. 

We observe that rigid classification significantly hinders performance, underscoring the need for trackers that effectively leverage multimodal training -- a promising yet underexplored area.

\begin{table}[t]
    \centering
        \caption{Rigid or Soft Classifier.}
        \vspace{-3mm}
    \resizebox{\columnwidth}{!}{
    \begin{tabular}{c|ccc|cc|cc}
        \hline
        \multirow{2}{*}{Prob.} & \multicolumn{3}{c|}{DepthTrack} & \multicolumn{2}{c|}{LasHeR} & \multicolumn{2}{c}{VisEvent} \\
        \cline{2-8}

        & F-score & Re & Pr & Pr & Sr & Pr & Sr \\
        \hline

        100\%  & 58.8 & 57.7 & 58.2 & 68.4 & 55.1 &  76.4 &60.2  \\
        80\%  & \textbf{61.5} & \textbf{62.0} & \textbf{61.8} & \textbf{69.1} & \textbf{55.7} &\textbf{77.5}   &  \textbf{60.9}\\
        33\%  & 61.0 & 60.6 & 60.8 & 69.0 & 55.4 & 77.0  & 60.5 \\
        \hline
    \end{tabular}}

    \vspace{-2mm}
    \label{tab:Ablation3}
\end{table}

\vspace{1mm}

\noindent \textbf{Visualization and Analysis:}
To better understand the routing mechanism, we provide in Fig. \ref{fig:router} the top-k choice visualization for different challenging scenes, all with RGB-Event only as input for inference. In our case, we set $k=2$:

\noindent - In Scene 1, a low-light scenario, thermal sensing contributes to lighting awareness. Visualizations show that alongside the primary event expert, several thermal experts are activated, enhancing feature modeling and adapting to low-light-induced appearance changes. 

\noindent - Scene 2 focuses on car tracking, where activated foreground events are mixed with background events. To better model geometry and facilitate foreground-background separation, our network employs depth experts. Thermal experts are also triggered here, likely due to overexposure.

\noindent - Scene 3 involves tracking with a heavily blurred image, our model primarily relies on event experts, aligning well with human intuition, as event cameras excel in such a task.

Additionally, we observe that the routing choice can be influenced by network depth, as deeper layers may handle different features compared to shallower layers.

\vspace{1mm}
\noindent \textbf{Limitations, Future Work, and Social Impact:} In this paper, we explore the potential of cross-modal knowledge transfer for multimodal tracking. A limitation of our current approach is the use of identical expert designs for each modality. We believe that some attributes may be more salient than others, which should naturally lead to an adaptive computational cost. However, this has not been thoroughly studied in multimodal tracking or more broadly in general multimodal tasks, and we aim to investigate this in future work. Finally, object tracking has widespread applications, and we advocate for its use exclusively in socially beneficial contexts.

\section{Conclusion}
In this paper, we introduce a novel approach to multimodal knowledge sharing for visual tracking, addressing the core challenge of data sparsity in multimodal sensing and the significant factors leading to appearance discrepancies. Our key innovation is a weak classifier-based mechanism that facilitates cross-modal knowledge transfer when samples from different modalities share similar characteristics. By integrating this mechanism within a mixture-of-experts framework, we enable effective cross-modal knowledge sharing. Our method successfully bridges the gap between multimodal training and single-modality deployment, offering a practical solution to the challenge of leveraging multimodal data effectively. The experimental results demonstrated the effectiveness of our approach, achieving a significant overall improvement over SOTA methods across all RGB-X tracking scenarios.